\documentclass[10pt,twocolumn,letterpaper]{article}

\usepackage{iccv}
\usepackage{times}
\usepackage{epsfig}
\usepackage{graphicx}
\usepackage{amsmath}
\usepackage{amssymb}
\usepackage{authblk}
\usepackage[normalem]{ulem}

\usepackage[breaklinks=true,bookmarks=false]{hyperref}

\iccvfinalcopy % *** Uncomment this line for the final submission

\ificcvfinal\fi
\begin{document}

%%%%%%%%% TITLE
\title{Efficient and Differentiable Shadow Computation for Inverse Problems}

\author{Linjie Lyu, Marc Habermann, Lingjie Liu, Mallikarjun B R,  Ayush Tewari, and Christian Theobalt}

\affil{Max Planck Institute for Informatics, Saarland Informatics Campus}


\maketitle

%%%%%%%%% ABSTRACT

\begin{abstract}
Differentiable rendering has received increasing interest for image-based inverse problems. 
It can benefit traditional optimization-based solutions to inverse problems, but also allows for self-supervision of learning-based approaches for which training data with ground truth annotation is hard to obtain.
However, existing differentiable renderers either do not model visibility of the light sources from the different points in the scene, responsible for shadows in the images, or are too slow for being used to train deep architectures over thousands of iterations.
To this end, we propose an accurate yet efficient approach for differentiable visibility and soft shadow computation. 
Our approach is based on the spherical harmonics approximations of the scene illumination and visibility, where the occluding surface is approximated with spheres.
This allows for a significantly more efficient shadow computation compared to methods based on ray tracing. 
As our formulation is differentiable, it can be used to solve inverse problems such as texture, illumination, rigid pose, and geometric deformation recovery from images using analysis-by-synthesis optimization.
\end{abstract}

%%%%%%%%% BODY TEXT

\begin{figure}
	\centering
	\includegraphics[width=0.8\linewidth]{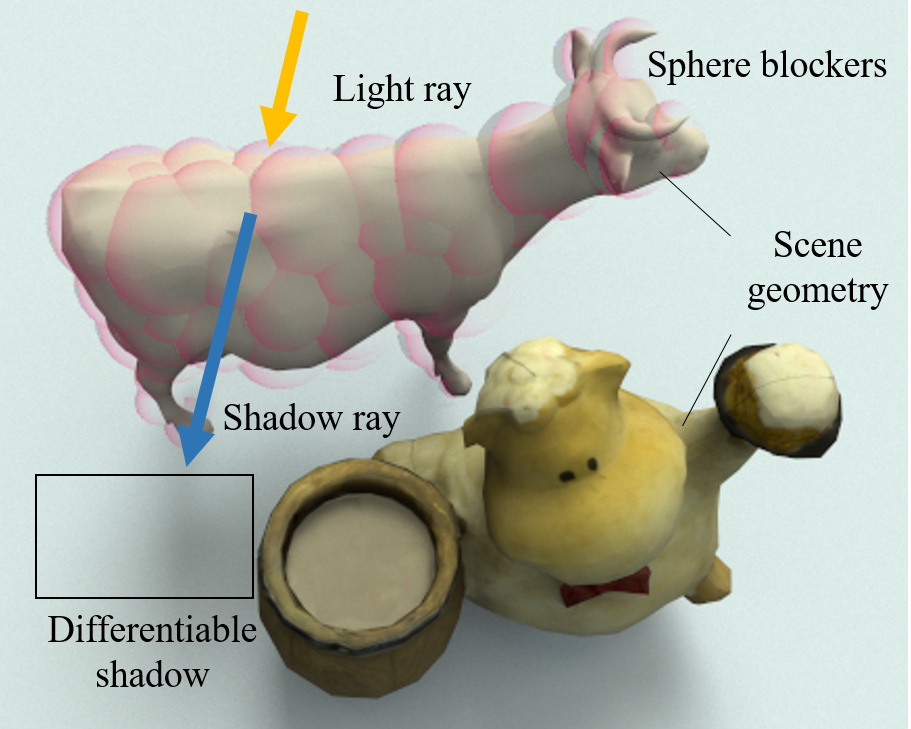} 
	\caption
	{
	    We propose a fast and differentiable visibility approximation for rendering which allows us to account for shadow effects.
	    To this end, we approximate the geometry with a sphere set and compute visibility in the spherical harmonics space.
	}
	\label{fig:teaser}
\end{figure}

%
%%%%%%%%%%%%%%%%%%%%%%%%%%%
%
\section{Introduction}
\label{sec:introduction}
%
%%%%%%%%%%%%%%%%%%%%%%%%%%%
%
% Motivation
%
Rendering virtual scenes, objects, and characters has a wide range of applications in movies, video games, and many other areas which require synthesis of realistic images.
While in these computer graphics applications the main interest is the generation of images from scene parameters like geometry, lighting, and texture, rendering can also be used to solve inverse problems, which attempt to recover exactly these scene parameters from real images.
Analysis-by-synthesis optimization is commonly used~\cite{thies2016face,Blanz} where the estimated scene parameters are rendered as a synthetic image and then compared with the reference. 
If the renderer is differentiable, an energy function which compares the renderings and the ground truth images can be used for optimization.
Differentiable rendering is also interesting for learning-based, notably neural network-based, approaches where ground truth annotations, e.g. of the dense geometry, are not easily available for large image-based training corpora.
Instead, differentiable rendering allows for self-supervised learning using an analysis-by-synthesis approach where the rendered image is compared to the real one.
This has been widely used in the vision and machine learning community, for solving problems such as reflectance estimation~\cite{azinovic2019inverse,DADDB18,li2020inverse}, free-viewpoint synthesis~\cite{thies2019deferred,lombardi2019neural}, and human performance capture~\cite{Habermann:2019:LRH:3313807.3311970,deepcap,tewari17MoFA,8496850}.
%
%%%%%%%%%%%%%%%%%%%%%%%%%%%
%
% rasterization based techniques
%
\par 
Most differentiable rendering methods rely on rasterization-based techniques which only consider direct illumination effects~\cite{kato2018renderer,ravi2020pytorch3d,liu2019soft,laine2020modular}. 
Shadows are global illumination effects, i.e., for any point in the scene, any other point could be occluding the light source, see Fig.~\ref{fig:teaser}.
Thus, they are not modeled by the direct illumination renderers. 
As a result, inverse methods supervised with such differentiable renderers produce undesired artifacts. 
For instance, the estimated texture, geometry, and illumination may exhibit baked in errors trying to represent real world effects, in particular due to shadows which are not accounted for by these simplified rendering assumptions.
In this paper, we address the problem of illumination visibility, i.e., whether the light source in a direction is visible from any point in the scene. 
Illumination visibility is simply called "visibility" for readability in the paper, not to be confused by camera visibility, i.e., which points of the scene are visible in the image.
%
%%%%%%%%%%%%%%%%%%%%%%%%%%%
%
% Global illuminatioon
%
\par 
To account for these limitations, differentiable ray tracing methods~\cite{li2018differentiable,10.1145/3355089.3356522} were proposed which use ray tracing methods to solve the rendering equation.
Some approaches~\cite{azinovic2019inverse} use them for reconstructing more accurate scene parameters compared to direct illumination-based techniques.
While these renderers can render shadows and other higher-order illumination effects, they are computationally inefficient, which makes training large networks practically impossible on consumer-grade hardware.
%
%%%%%%%%%%%%%%%%%%%%%%%%%%%
%
% our method
%
\par 
To this end, we propose a method for differentiable and efficient visibility computation for rendering scenes with soft-shadows.
Our work builds up on the literature of efficient global rendering~\cite{ren2006real,sloan2002precomputed,Guerrero08} where the goal is to approximate visibility for a faster runtime.
More precisely, our approach first approximates the scene geometry with spheres. 
These spheres are attached/rigged to the underlying geometry mesh, which allows for deforming and posing the mesh through the sphere representation.
Scene illumination is modeled with the commonly used spherical harmonics representation~\cite{10.1145/383259.383317}.
Interestingly, the same representation can also be used to model visibility, e.g. whether the incident illumination is occluded in any direction from a point in the scene.
This spherical harmonics representation allows for efficient rendering of soft shadows using fast spherical harmonics multiplications. 
We combine this soft shadow rendering with a diffuse spherical harmonics based shading model to obtain the final rendering which is fully differentiable enabling us to compute gradients with respect to geometry, light, and texture.
We show applications of this differentiable renderer, by using analysis-by-synthesis optimization in order to recover the rigid pose, surface deformation, scene illumination, and texture of objects in scenes with shadows.
%
%%%%%%%%%%%%%%%%%%%%%%%%%%%
%
% Summary
%
In summary, our contributions are:
\begin{itemize}
\item{A differentiable and efficient renderer which can synthesize soft-shadows for dynamic scenes.}
\item{The integration of our renderer in an optimization-based setup for the reconstruction of scene parameters from monocular images.}
\end{itemize}
We compare our approach to the state-of-the-art differentiable rendering techniques and show that our method offers a good trade-off between rasterization-based techniques which are efficient but do not model shadows, and the more accurate but inefficient ray tracing approaches.
We will make our implementation publicly available.
\section{Related Work}
\label{sec:relatedwork}
%
%%%%%%%%%%%%%%%%%%%%%%%%%%%%%%%%%
%
Differentiable rendering is a widely studied problem.
In this section, we will discuss the most relevant methods. 
We refer to Kato~\etal~\cite{kato2018renderer} for a recent detailed survey.
Existing differentiable rendering methods are either based on efficient but inaccurate direct illumination or more accurate but inefficient global illumination. 
%
%%%%%%%%%%%%%%%%%%%%%%%%%%%%%%%%%%%%%%%%%%%%%%%%%%%%%%%%%%%%%%%%%%%%%%%%%%%%
%
\subsection{Differentiable Rendering}
Efficient but approximate differentiable approaches can be further split into two categories based on the type of approximations. 
Some works~\cite{Loper:ECCV:2014,kato2019vpl, kato2018renderer,laine2020modular} approximate gradients without modifying the rasterization step which has the advantage that camera visibility is modeled as in the real world.
However, the camera visibility computation here is non-differentiable.
In contrast, there are other works~\cite{Rhodin:2015, liu2019soft} that treat the objects in the scene as semi-transparent volumes.
This allows for differentiable camera visibility but their method does not reflect the real world properties of the object. 
All of the above methods do not account for the visibility of the light sources and hence cannot account for cast shadows and self-shadows. 
This leads to inaccurate results, e.g. the geometry can deform incorrectly to explain shadows in the image, or the recovered textures can contain baked in shadows.
In contrast, our proposed approximation to global illumination results in meaningful supervision even in the presence of shadows as we explicitly model them.
\par
Recently, physically-based differentiable rendering methods were proposed, which can also account for global illumination. 
These methods build up on Monte Carlo ray tracing which provides derivatives for arbitrary bounces of light. 
Li~\etal~\cite{li2018differentiable} proposed the first comprehensive method which can provide gradients for all scene parameters.
Zhang~\etal~\cite{10.1145/3355089.3356522} proposed a similar approach which also accounts for volumetric derivatives along with meshes.
Please refer to Zhao~\etal~\cite{pbr_tut} for a detailed analysis of the various physically-based differentiable rendering methods.
While these methods provide accurate supervision with respect to global illumination effects, they are very slow to evaluate as for a single pixel many rays have to be sampled and accumulated. 
This makes it difficult or nearly impossible to use them within the training of neural networks.
%
%%%%%%%%%%%%%%%%%%%%%%%%%%%%%%%%%
%
\begin{figure*}[t]
\centering
\includegraphics[width=0.9\textwidth]{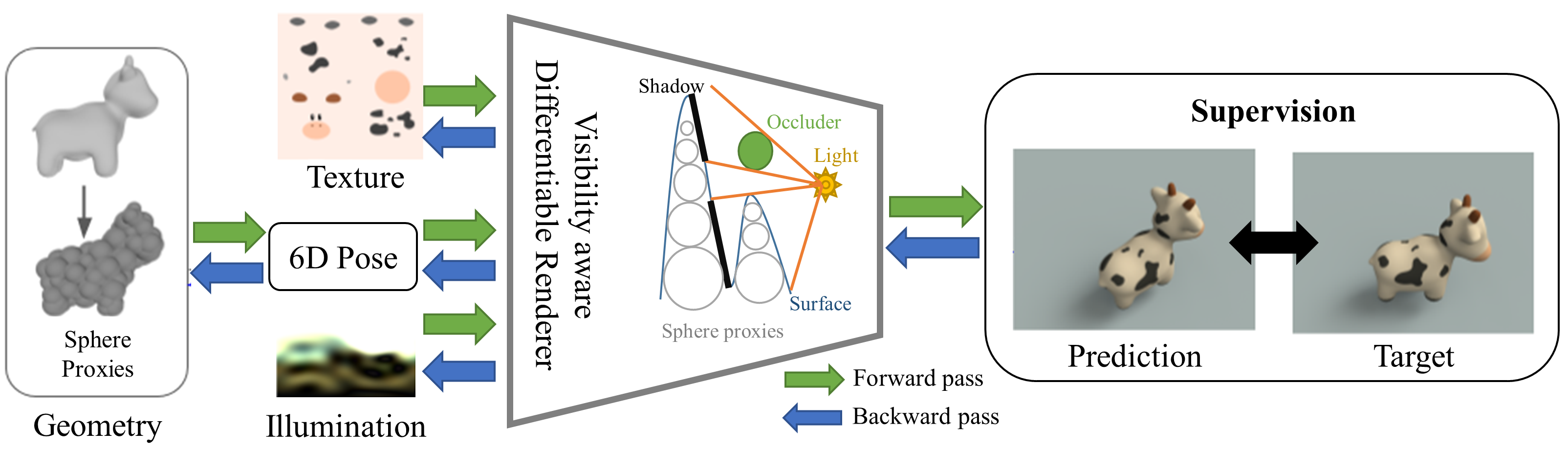} 
\caption
{
Overview of our approach. 
Given a surface mesh, we first approximate the geometry surface with a set of spheres.
The global visibility can be calculated as a combination of the visibility function for each sphere blocker in the spherical harmonics space, where the function is associated with texture, pose and illumination.
Combined with a rasterizer, an image can be rendered in a differentiable way. 
Therefore, we are able to optimize the different scene properties, such as geometry, texture, and illumination by comparing the rendered image against the target image.
}
\label{fig:overview}
\end{figure*}
%
%%%%%%%%%%%%%%%%%%%%%%%%%%%%%%%%%%%%%%%%%%%%%%%%%%%%%%%%%%%%%%%%%%%%%%%%%%%%
%
\subsection{Efficient Global Illumination}
Several methods in the computer graphics literature have explored faster global illumination approaches, see the survey of Ritschel~\etal~\cite{ritschel2012state}.
Here, precomputed Radiance Transfer (PRT) methods are related to our approach. 
Most PRT methods~\cite{dobashi1995quick,sloan2002precomputed,kautz2002fast} assume the geometry to be fixed, although some methods work with dynamic geometry~\cite{sloan2005local,iwasaki2007precomputed}. 
Very recently, PRT was also used for inverse problems~\cite{thul2020precomputed}. 
However, the scene geometry cannot be updated in this formulation.
Several approaches have been proposed for efficient computation of soft-shadows for dynamic scenes~\cite{kautz2004hemispherical,zhou2005precomputed,kontkanen2005ambient}.
Ren~\etal ~\cite{ren2006real} used spherical harmonics (SH) representations for the different scene components such as illumination and visibility and proposed an efficient, but non-differentiable method for computing SH products.
Zhou ~\etal ~\cite{zhou2005precomputed} proposed shadow fields to represent the light source radiance and occlusions, which allows for fast computation of the visibility. 
Efficient soft-shadow computation has mostly been explored in computer graphics for creating synthetic imagery. 
We investigate the inverse problem: Using an efficient \emph{differentiable} renderer for estimating scene parameters from monocular images.
%
%
%%%%%%%%%%%%%%%%%%%%%%%%%%%%%%%%%%%%%%%%%%%%%%%%%%%%%%%%%%%%%%%%%%%%%%%%%%%%%%%
%
\section{Method}
\label{sec:method}
Our method solves an efficient approximation of the rendering equation for generating soft shadows for diffuse surface meshes.
Traditional graphics rendering is extremely time-consuming, mainly due to the expensive sampling process used in ray tracing for computing global illumination. 
To approach this problem, we represent the geometry surface with a collection of sphere blockers and compute the global visibility as a combination of the visibility function for each sphere blocker (also referred to as blocker function) in the spherical harmonics (SH) space, see Fig.~\ref{fig:overview}. 
This computation in the SH space is efficient and differentiable, enabling fast rendering as well as optimization of the scene parameters. 
Our approach is closely related to the work of Ren~\etal~\cite{ren2006real} which uses a similar formulation. 
However, we provide a novel algorithmic formulation, which allows differentiating through the rendering process.
Importantly, we show, for the first time, solutions of inverse problems with global effects using an efficient differentiable renderer. 
In the following, we will first introduce the rendering equation (Sec.~\ref{render_equ}) and its spherical harmonics approximations (Sec.~\ref{sec:methodSphericalHarmonics}) as our method builds up on these concepts.
We will then describe the rendering process of our approach (Sec.~\ref{sec:SH_compute}-\ref{diffuse_shading}) and the optimization of the scene properties with our differentiable renderer (Sec.~\ref{sec:IBO}). 
%
%%%%%%%%%%%%%%%%%%%%%%%%%%%%%%%%%%%%%%%%%%%%%%%%%%%%%%%%%%%%%%%%%%%%%%%%%%%%%%%
%
\subsection{Rendering Equation}
\label{render_equ}
The rendering equation~\cite{10.1145/15886.15902} with only diffuse non-emitting surfaces in the scene is defined as:
%
%%%%%%%%%%%%%%%%%%%%%%%%%%%%%%%%%
%
\begin{equation}
\label{eq:renderingEquation}
B(\mathbf{x})=a(\mathbf{x}) \int_{\Omega} L(\omega,\mathbf{x}) max((\omega\cdot\mathbf{n}(\mathbf{x})),0) d\omega\,,
\end{equation}
%
%%%%%%%%%%%%%%%%%%%%%%%%%%%%%%%%%
%
where $B(\mathbf{x})$ is the outgoing radiance at point $\mathbf{x}$, $L(\omega,\mathbf{x})$ is the incoming radiance at this point from direction $\omega$, $\Omega$ represents the sphere of directions, $a(\mathbf{x})$ is the diffuse albedo, and $\mathbf{n}(\mathbf{x})$ is the normal at point $\mathbf{x}$.
Assuming no inter-reflectance, the incoming radiance $L(\omega,\mathbf{x})$ at point $x$ can be decomposed into the static environment lighting $L(\omega)$ and the visibility $V(\omega,\mathbf{x})$ at point $x$. 
Note that the environment light is only a function of the light direction while the visibility depends on both the light direction and the point location, since it is a global illumination property. 
Given this, we can rewrite the rendering equation as:
%
%%%%%%%%%%%%%%%%%%%%%%%%%%%%%%%%%
%
\begin{equation}
\label{eq:render}
\begin{split}
B(\mathbf{x})&= a(\mathbf{x})\int_{\Omega} L(\omega) V(\omega,\mathbf{x}) H(\omega, \mathbf{x}) d\omega \enspace{,} 
\end{split}
\end{equation}
%
%%%%%%%%%%%%%%%%%%%%%%%%%%%%%%%%%
%
where $H(\omega, \mathbf{x}) = max((\omega\cdot\mathbf{n}(\mathbf{x})),0)$.
One way to compute $B(\mathbf{x})$ is with Monte Carlo integration, which is computationally expensive as lots of samples have to be acquired to approximately evaluate the integral.
Instead, we estimate a fast approximation of this integration by using spherical harmonics as introduced next. 
%
%%%%%%%%%%%%%%%%%%%%%%%%%%%%%%%%%%%%%%%%%%%%%%%%%%%%%%%%%%%%%%%%%%%%%%%%%%%%%%%
%
\subsection{Spherical Harmonics Approximation}
\label{sec:methodSphericalHarmonics}
A spherical function can be projected into spherical harmonics space and reconstructed back using the spherical harmonics bases. 
We will write variables corresponding to SH coefficients in bold letters in the following. 
When representing a function with a subset of basis functions within a limited bandwidth, we obtain a low frequency approximation of the original function. 
In detail, given a spherical function $f(\omega)$, its corresponding SH vector $\boldsymbol{f}$ (with all SH coefficients stacked) is defined as
%
%%%%%%%%%%%%%%%%%%%%%%%%%%%%%%%%%
%
\begin{equation}
\label{eq:coes}
\boldsymbol{f}_i=\int_{\Omega} f(\omega) y_i(\omega) d \omega ,
\end{equation} 
%
%%%%%%%%%%%%%%%%%%%%%%%%%%%%%%%%%
%
where $y_i(\omega)$ are the SH basis functions.
The indices are linearized, with $i$ = ${l^{2}+l+m}$, where $l$ is the index of the SH band, and $m$, $-l\leq m \leq l$ is the index within the band. 
Computing Eq.~\ref{eq:coes} requires solving the integral which is usually achieved by Monte Carlo integration. 
For an inverse problem, this can be very expensive for both forward and backward computations.
As a remedy, we precompute several terms which allows for efficient and differentiable computation as we show later. 
Given the SH coefficients and the basis functions, the original signal can be approximated as 
%
%%%%%%%%%%%%%%%%%%%%%%%%%%%%%%%%%
%
\begin{equation}
\Tilde{f}(\omega) = \sum_{i=1}^{n^2} \boldsymbol{f}_i y_i(\omega) ,
\end{equation}
%
%%%%%%%%%%%%%%%%%%%%%%%%%%%%%%%%%
%
where $i$ is the linearized index explained above, and $n$ is the number of bands chosen to approximate the signal. 
The accuracy of reconstruction increases with the number of bands. 
In our case, we choose a bandwidth of $n$ = 8, which we found to be good for low-frequency approximations.
The integral of a multiplication of two SH functions can be simply computed as a dot product of their coefficients, 
%
%%%%%%%%%%%%%%%%%%%%%%%%%%%%%%%%%
%
\begin{equation}
\int_{\Omega} \tilde{a}(\omega) \tilde{b}(\omega) d\omega = \mathbf{a} \cdot \mathbf{b} \,.
\end{equation}
%
%%%%%%%%%%%%%%%%%%%%%%%%%%%%%%%%%
%
The SH projection of this multiplication is computed as 
%
%%%%%%%%%%%%%%%%%%%%%%%%%%%%%%%%%
%
\begin{align}
\label{eq:SH_mult}
\mathbf{a} * \mathbf{b} = \int_{\Omega} \tilde{a}(\omega) \tilde{b}(\omega) y(\omega) d\omega , \\
(\mathbf{a} * \mathbf{b})_i = \sum_{j=1}^{n} \sum_{k=1}^{n} C_{ijk} \mathbf{a_j} \mathbf{b_k} , 
\end{align}
%
%%%%%%%%%%%%%%%%%%%%%%%%%%%%%%%%%
%
where $C_{ijk} = \int_{\Omega} y_i(\omega) y_j(\omega) y_k(\omega)$ is a transfer matrix which can be precomputed. 
Please refer to Green~\cite{green2003spherical} for the \emph{gritty details}.
In our case, we first compute the SH coefficients for $L$, $V$ and $H$ in Eq.~\ref{eq:render}. 
With that, the solution of $B(\mathbf{x})$ can be analytically computed as 
%
%%%%%%%%%%%%%%%%%%%%%%%%%%%%%%%%%
%
\begin{equation}
B(\mathbf{x}) = a(\mathbf{x}) \Big(\boldsymbol{L} \cdot (\boldsymbol{V}(\mathbf{x}) * \boldsymbol{H}(\mathbf{x})) \Big) \enspace{.}
\end{equation}
%
%%%%%%%%%%%%%%%%%%%%%%%%%%%%%%%%%
%
This way of computing the outgoing radiance does not require inefficient sampling-based integration, which can also be sensitive to the sampling strategy used. 
Next, the efficient computation of these SH coefficients is explained. 
%
%%%%%%%%%%%%%%%%%%%%%%%%%%%%%%%%%%%%%%%%%%%%%%%%%%%%%%%%%%%%%%%%%%%%%%%%%%%%%%%
%
\subsection{Spherical Harmonics Computations}
\label{sec:SH_compute}
We first compute $\boldsymbol{L}$ and $\boldsymbol{H(\mathbf{x})}$ as 
%
%%%%%%%%%%%%%%%%%%%%%%%%%%%%%%%%%
%
\begin{equation}
\label{eq:light_SH}
\boldsymbol{L}= \int_{\Omega} L(\omega) {y}(\omega) d \omega \,,
\end{equation} 
%
%%%%%%%%%%%%%%%%%%%%%%%%%%%%%%%%%
%
\begin{equation}
\label{eq:SH_H}
\boldsymbol{H(\mathbf{x})}= \int_{\Omega} max((\mathbf{n(\mathbf{x})} \cdot \omega),0) {y}(\omega) d \omega \,.
\end{equation} 
%
%%%%%%%%%%%%%%%%%%%%%%%%%%%%%%%%%
%
We solve Eq.~\ref{eq:light_SH} using numerical integration at initialization. 
This equation does not have to be differentiable, since we only want to update the SH coefficients of the environment map.
For Eq.~\ref{eq:SH_H}, we formulate the computation of $max((\mathbf{n(\mathbf{x})} \cdot \omega),0)$ as 
%
%%%%%%%%%%%%%%%%%%%%%%%%%%%%%%%%%
%
\begin{equation}
\label{eq:reparameterize}
\boldsymbol{H(\mathbf{x})}= \phi_\text{SH}(\mathbf{n(x)}) \int_{\Omega} max(i_\mathrm{z} \cdot \omega,0) \boldsymbol{y}(\omega) d \omega \,.
\end{equation} 
%
%%%%%%%%%%%%%%%%%%%%%%%%%%%%%%%%%
%
Here, $\phi_\text{SH}(\mathbf{x})(\cdot)$ is a linear function which defines the rotation of spherical harmonic functions~\cite{kautz2002fast} and $i_\mathrm{z}=(0,0,1)^\top$. 
This reformulation has several advantages. 
First, this makes the equation differentiable with respect to $\mathbf{n(x)}$.
Second, spherical harmonics projections of functions which are symmetric along the z-axis can be computed analytically and efficiently. 
This leaves us with the visibility term.
%
%%%%%%%%%%%%%%%%%%%%%%%%%%%%%%%%%%%%%%%%%%%%%%%%%%%%%%%%%%%%%%%%%%%%%%%%%%%%%%%
%
\subsubsection{Sphere Fitting and Geometry Deformation}
\label{spherefitting}
Our method approximates the geometry of the non-deformed initial mesh with a set of spheres for computing the visibility, see Fig.~\ref{fig:overview}.
The mesh can then be deformed by translating and rotating these spheres. 
We can thus also update the geometry of the mesh using our renderer.
Visibility related rendering effects like shadows will supervise the sphere parameters and therefore the underlying geometry. 
To determine the initial position and radius of each sphere for a given geometry, we minimize the total volume occupied by the sphere set which is outside the mesh, referred to as sphere outside volume ~\cite{wang2006variational} along with a term encouraging the sphere set to cover as much of the mesh surface as possible. 
This objective is optimized using gradient descent. 
Please refer to the supplemental document for more details.
We then connect the sphere centers as an embedded graph to drive the deformation of the mesh geometry~\cite{sumner2007embedded} where each sphere is connected to its K-nearest vertices on the template. 
The mesh deformation is parameterized by the rotation and translation of each sphere. 
Later, the inverse problems optimize the rotation and translation of the spheres using the rendering loss (see Eq.~\ref{eq:render_loss}).
%
%%%%%%%%%%%%%%%%%%%%%%%%%%%%%%%%%%%%%%%%%%%%%%%%%%%%%%%%%%%%%%%%%%%%%%%%%%%%%%%
%
\subsubsection{Visibility in Spherical Harmonics Space}
Similar to Zhou~\etal~\cite{zhou2005precomputed}, we calculate $\mathbf{V}(\mathbf{x})$ as the product of the SH vectors of all the sphere blockers:
%
%%%%%%%%%%%%%%%%%%%%%%%%%%%%%%%%%
%
\begin{equation}
\label{eq:visibility_multiplication}
\boldsymbol{V}(\mathbf{x})=\boldsymbol{V}_{1}(\mathbf{x}) * \boldsymbol{V}_{2} (\mathbf{x})* \ldots * \boldsymbol{V}_{n}(\mathbf{x}) \,.
\end{equation}
%
%%%%%%%%%%%%%%%%%%%%%%%%%%%%%%%%%
%
where $\boldsymbol{V}_{i}(\mathbf{x})$ is the SH vector of the blocker function of the sphere $S_{i}$, which measures the blocking effect of a sphere:
%
%%%%%%%%%%%%%%%%%%%%%%%%%%%%%%%%%
%
$$
V_{i}(\omega,\mathbf{x})=\left\{\begin{array}{ll}
0 \text {,} & \text{ if } S_i \text { blocks light in direction } \omega \text{;} \\
1 \text{,} & \text{ otherwise. }
\end{array}\right. \,.
$$
%
%%%%%%%%%%%%%%%%%%%%%%%%%%%%%%%%%
%
Similar to Eq.~\ref{eq:reparameterize}, the computation of $\boldsymbol{V}_i(\mathbf{x})$ can be reparameterized to be differentiable with respect to $\mathbf{x}$.
The higher-order product in Eq.~\ref{eq:visibility_multiplication} can be computed using a series of SH multiplications as defined in Eq.~\ref{eq:SH_mult}. 
However, this can be computationally expensive. 
To accelerate this computation, we adopt the method proposed by Ren~\etal~\cite{ren2006real}, where the exponential of logarithm of SH functions are computed instead. 
The logarithm of the product leads to a summation, which can be computed efficiently. 
We follow their scaling, squaring, and optimal linear approximation setting.
Ren~\etal~\cite{ren2006real} compute the logarithm of the SH functions based on a lookup table which is not differentiable with respect to the sphere orientations.
Instead, we propose a differentiable approach which works well for our setting. 
We first approximate $V_i$ as $V^{'}_{i}(\omega,\mathbf{x})$: 
%
%%%%%%%%%%%%%%%%%%%%%%%%%%%%%%%%%
%
\begin{equation}
V^{'}_{i}(\omega,\mathbf{x})=\left\{\begin{array}{ll}
e^{-\epsilon} \text {,} & \text{ if } S_i \text { blocks in direction } \omega \text{:} \\
1 \text {,} & \text{ otherwise. }
\end{array}\right. 
\end{equation}
%
%%%%%%%%%%%%%%%%%%%%%%%%%%%%%%%%%
%
$\epsilon$ is set to $3$ such that $e^{-3} \approx 0.05$. 
This avoids the infinite logarithm for 0. 
We orient $\log (V^{'}_{i}(\omega,\mathbf{x}))$, the logarithm of this approximated function for each sphere, to the $z$ axis and project it to SH space. 
This computation has an analytical form with respect to the distance to the sphere and its radius.
Then, we apply the SH rotation as in Eq.~\ref{eq:reparameterize} with the vector pointing to the sphere center from $\mathbf{x}$. 
Finally, we add them and exponentiate the result to compute $\boldsymbol{V}(\mathbf{x})$.
Please refer to the supplemental for more details. 
%
%%%%%%%%%%%%%%%%%%%%%%%%%%%%%%%%%%%%%%%%%%%%%%%%%%%%%%%%%%%%%%%%%%%%%%%%%%%%
%
\subsection{Rendering}
\label{diffuse_shading}
Combined with a rasterizer, we can render an image as
%
%%%%%%%%%%%%%%%%%%%%%%%%%%%%%%%%%
%
\begin{equation}
I_R=R( B(\mathbf{x}_{0}),\dots,B(\mathbf{x}_{i}),\dots,B(\mathbf{x}_{n}), P)
\end{equation}
%
%%%%%%%%%%%%%%%%%%%%%%%%%%%%%%%%%
%
where $I_R$ is the image intensity, $B(\mathbf{x}_{i})$ is the radiance of vertex $\mathbf{x}_{i}$ as computed in Eq.~\ref{eq:render}, $P$ is a projection matrix implementing the camera using its intrinsics and extrinsics and $R$ is the rasterization function. 
The radiance $B(\cdot)$ includes shadows as well as diffuse shading of the surface. 
%
%%%%%%%%%%%%%%%%%%%%%%%%%%%%%%%%%%%%%%%%%%%%%%%%%%%%%%%%%%%%%%%%%%%%%%%%%%%%
%
\subsection{Image-Based Optimization}
\label{sec:IBO}
We optimize the different scene properties, such as geometry, albedo, and illumination by comparing the rendered image to the reference image. 
In all experiments, we use the $\ell_2$ loss function between the rendered and reference image for optimization. 
The objective function can be written as:
%
%%%%%%%%%%%%%%%%%%%%%%%%%%%%%%%%%
%
\begin{align}
\label{eq:render_loss}
& \mathcal{L}(\theta, \boldsymbol{L}, a_{0,\dots,n}) = || I_R(\theta, \boldsymbol{L}, a_{0,\dots,n}) - I||^2_2 \,, \\
& I_R(\theta, \boldsymbol{L}, a_{0,\dots,n}) = R(B(\theta \mathbf{x_0}), \dots, B(\theta \mathbf{x_n}), P) \,.
\end{align}
%
%%%%%%%%%%%%%%%%%%%%%%%%%%%%%%%%%
%
Here, $\theta$ includes the global rigid pose and embedded deformation \cite{sumner2007embedded} of an object in the scene and $I$ is the reference image.
We can also optimize for a texture map by re-parameterizing $a$ in 2D as a pre-process. 
If there are multiple objects in the scene, they can have different rigid poses.
This loss function is differentiable since we use a differentiable rasterizer~\cite{ravi2020pytorch3d} as $R(\cdot)$, and our radiance computation $B(\mathbf{x})$ is differentiable with respect to $\theta$, $\boldsymbol{L}$ and $a_{0,..,n}$ .
We use gradient descent with a step size of ${10^{-2}}$ for optimization with $200-1000$ iterations.
%
%
%%%%%%%%%%%%%%%%%%%%%%%%%%%%%%%%%%%%%%%%%%%%%%%%%%%%%%%%%%%%%%%%%%%%%%%%%%%%%%%
%%%%%%%%%%%%%%%%%%%%%%%%%%%%%%%%%%%%%%%%%%%%%%%%%%%%%%%%%%%%%%%%%%%%%%%%%%%%%%%
%
\section{Results}
\label{sec:results}
\begin{figure}
	\centering
	\includegraphics[width=0.8\linewidth]{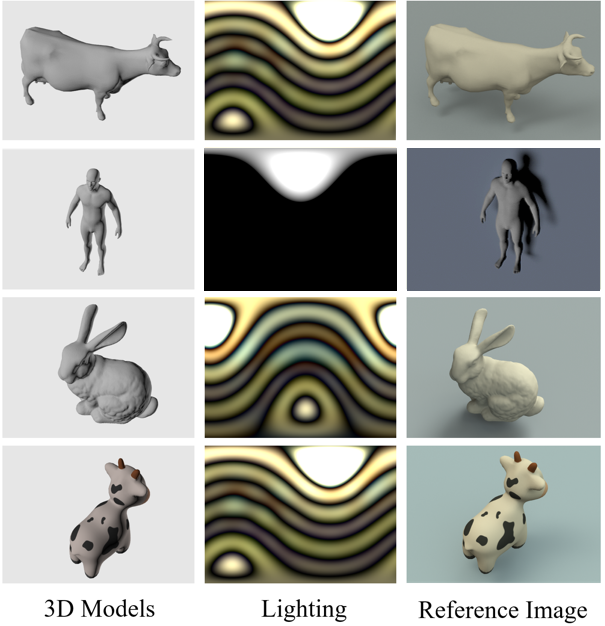} 
	\caption
	{
		Example scenes of our evaluation dataset.
		Note that geometry as well as lighting conditions are very complex making it a challenge for inverse problems.
	}
	\label{fig:scenes}
\end{figure}

Our method is implemented in Pytorch~\cite{NEURIPS2019_9015}.
We use an Intel(R) Xeon(R) Gold 6144 CPU processing unit and a Nvidia V100 graphics card for all results.
The supplemental video includes more visual results and comparisons.
%
%%%%%%%%%%%%%%%%%%%%%%%%%%%%%%%%%
%%%%%%%%%%%%%%%%%%%%%%%%%%%%%%%%%
%
\paragraph{Evaluation Dataset.}
To evaluate our approach, we create different virtual scenes containing various challenging geometries, textures, and lighting conditions.
In total, we use 7 different mesh models, e.g. animals and humans.
We acquired 10 environments maps~\cite{laval_hdr} containing various lighting conditions, e.g. outdoor skies.
These environment maps are projected onto the SH space.
We use a ray tracing approach~\cite{li2018differentiable} to render the reference images, where we choose 1024 sample rays per pixel to obtain a noise free reference.
Fig.~\ref{fig:scenes} shows some examples.
%
%%%%%%%%%%%%%%%%%%%%%%%%%%%%%%%%%
%
%
%%%%%%%%%%%%%%%%%%%%%%%%%%%%%%%%%
%
%%%%%%%%%%%%%%%%%%%%%%%%%%%%%%%%%
%%%%%%%%%%%%%%%%%%%%%%%%%%%%%%%%%
%
\subsection{Inverse Problems}
We conduct several experiments where the individual scene parameters are reconstructed from monocular images using analysis-by-synthesis optimization (Sec.~\ref{sec:IBO}). 
We compare to Redner~\cite{li2018differentiable}, which models global illumination using ray tracing, and to the implementation of Ravi~\etal~\cite{ravi2020pytorch3d} of a rasterization-based direct illumination (DI) method~\cite{10.1145/383259.383317}, which uses SH shading without shadows.
For Redner, we set the number of rays per pixel to 64 for all experiments, which was the smallest number that still gave us noise free renderings.
We use single bounce, as we do not evaluate indirect illumination effects.

\begin{figure*}
\includegraphics[width=\textwidth]{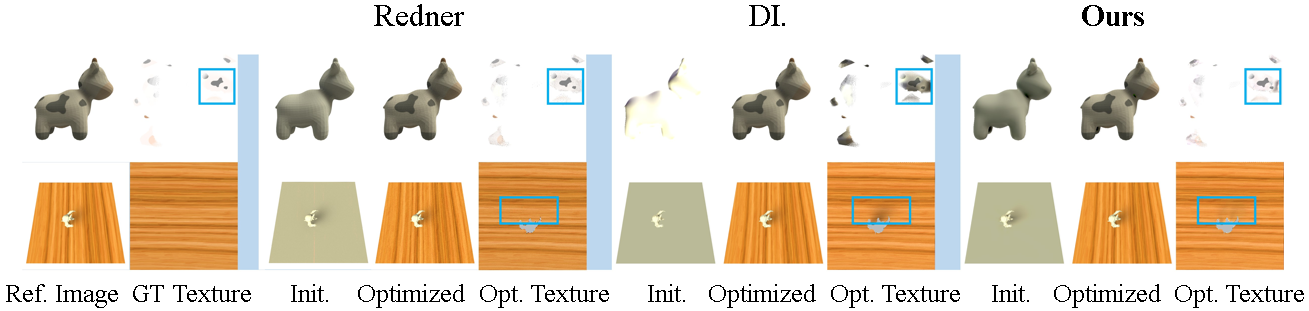} 
\caption
{
Texture optimization results.
From left to right. 
Ground truth rendering and texture map. 
Rendering with initial and optimized texture map as well as the optimized texture map for Redner, DI and our method.
Note that our approach outperforms DI method as they cannot remove the shadow in the texture and we are also close to Redner~\cite{li2018differentiable} while being much faster.
}
\label{fig:texResult}
\end{figure*}

\begin{figure*}
	\includegraphics[width=\linewidth]{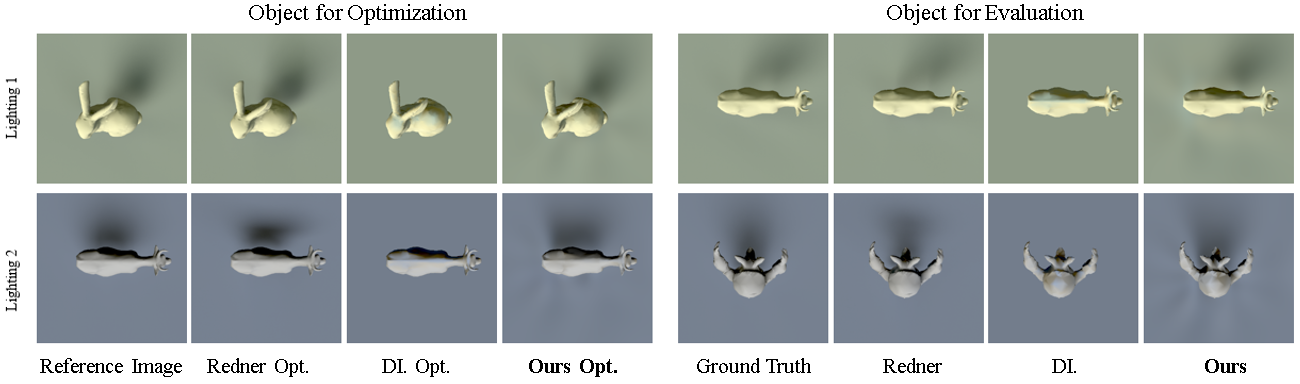} 
	\caption
	{
		Lighting optimization results. 
		From left to right. 
		The reference image used to optimize the scene lighting.
	    The rendered scene with the optimized scene lighting.
        The ground truth image for a new object.
        The rendering of the new object using the previously optimized scene lighting.
	    Note that our approach gives a good compromise between runtime and accuracy, compared to other approaches.
	}
	\label{fig:lightResult}
\end{figure*}

\begin{figure}
	\includegraphics[width=\linewidth]{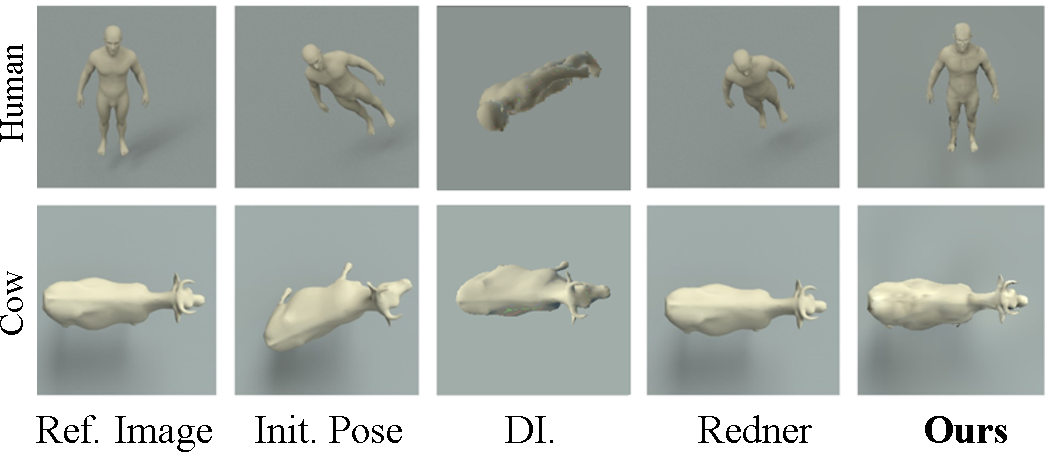} 
	\caption
	{
		6D Pose optimization result. 
		We outperform both direct illumination and global illumination methods.
	}
	\label{fig:Pose}
\end{figure}

\begin{figure}
	\includegraphics[width=\linewidth]{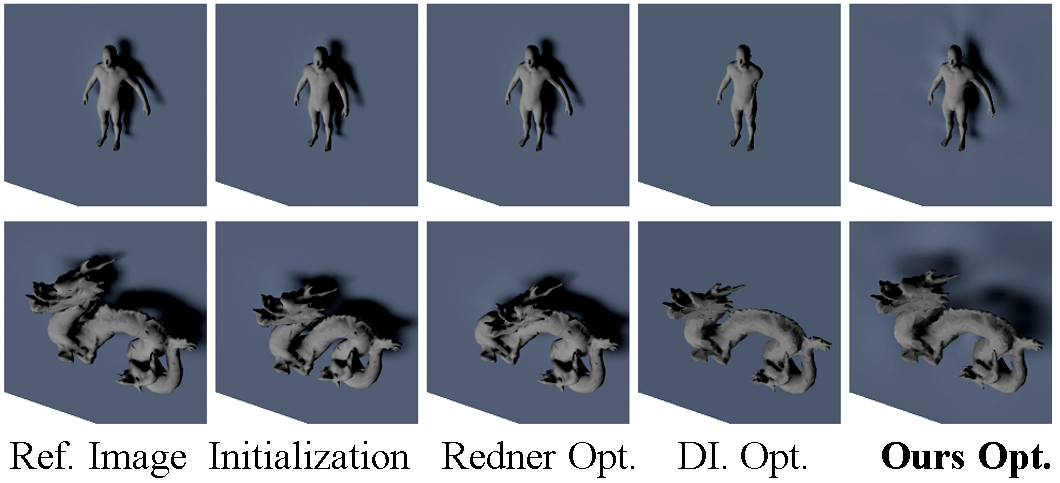} 
	\caption
	{
		Geometry optimization results. 
		We outperform both direct illumination and global illumination methods.
	}
	\label{fig:geometry}
\end{figure}

%%%%%%%%%%%%%%%%%%%%%%%%%%%%%%%%%
%%%%%%%%%%%%%%%%%%%%%%%%%%%%%%%%%
%s
\paragraph{Texture Optimization.}
\label{sec:texOpt}
First, we evaluate our approach for the purpose of texture estimation.
Given the geometry, lighting, and one reference image of the scene, we optimize for the diffuse albedo texture, initialized as pure white, using the different rendering methods.
Note that the reference images shown in Fig.~\ref{fig:texResult} contain shadows cast by the occluders, as well as self-shadows which makes it particularly challenging to recover the correct albedo.
In the first row, the texture of the cow is optimized.
However, a large sphere (not visible in the image) is also placed in the scene blocking the light coming from a large area of the environment map.
In the second row, the texture of the ground plane should be recovered onto which the armadillo is casting a shadow.
In both cases, our approach recovers albedo which is almost free from shadows, due to our shadow-aware renderer.
Note that DI method~\cite{ravi2020pytorch3d} cannot account for the shadow which manifests in the bright initialization (top row), as the occluding sphere is ignored and all the light arrives at the surface of the cow.
More importantly, these methods bake the shadows into the texture. 
In contrast, Redner~\cite{li2018differentiable} also obtains shadow-free textures but at a significantly slower speed.
This is also quantitatively evaluated in Tab.~\ref{tab:texture} where we measure the mean squared pixel error (MSE) between the optimized and the ground truth texture maps for all pixels that are visible in the rendered image.
Although Redner performs slightly better, our approach is very close in quality while being much faster.
Our method clearly outperforms the direct illumination renderer. 
%
%%%%%%%%%%%%%%%%%%%%%%%%%%%%%%%%%
%
%
%%%%%%%%%%%%%%%%%%%%%%%%%%%%%%%%%
%
%%%%%%%%%%%%%%%%%%%%%%%%%%%%%%%%%
%
\begin{table}
	\begin{center}
		\begin{tabular}{|c|c|}
			\hline
			\multicolumn{2}{|c|}{\textit{Texture Reconstruction Accuracy}} \\
			\hline
			\textbf{Method}                                             & \textbf{MSE}$\downarrow$     	    \\
			\hline
			Redner~\cite{li2018differentiable}							& \textbf{0.0213}		           	\\
			DI~\cite{ravi2020pytorch3d} 			                    & 0.2763		           			\\
			\hline
			Ours    									   	            & 0.0272		  				    \\
			\hline
		\end{tabular}
	\end{center}
	\caption
	{
		Texture reconstruction accuracy averaged over 10 scenes.
		We quantitatively outperform DI method  using mean squared error, as shadows are  baked in the texture for these approaches.
	}
	\label{tab:texture}
\end{table}
%
%%%%%%%%%%%%%%%%%%%%%%%%%%%%%%%%%
%
%%%%%%%%%%%%%%%%%%%%%%%%%%%%%%%%%
%%%%%%%%%%%%%%%%%%%%%%%%%%%%%%%%%
%
\paragraph{Lighting Optimization.}
Next, we evaluate our approach in terms of lighting reconstruction.
Here the geometry and albedo texture of the object are known, and we are interested in reconstructing the lighting from a reference image.
The reference images are shown in the left column of Fig.~\ref{fig:lightResult}.
We initialize all methods with no lights, i.e. all SH coefficients are set to zero.
The converged results are shown in Fig.~\ref{fig:lightResult} for two different scenes.
Moreover, we relight new objects with the optimized scene lighting, and compare it to the ground truth for an additional evaluation of the estimate.
We do not directly compute quantitative errors on the estimated environment maps because it is not easy to compute the visible regions in the environment map. 
Computing the error on the full environment map would not be indicative of the quality, as the occluded regions could be arbitrarily different for each method. 
Our indirect metric using a different object does not face this issue.
We outperform the DI method, as they cannot model the shadows.
As expected, the ray tracing method gives the most accurate results at the cost of a slow runtime.
In Tab.~\ref{tab:lighting}, we further quantitatively evaluate the accuracy of the lighting estimation \textit{on the new objects} in terms of mean squared error (MSE) by comparing the scene rendering with the ground truth.
Our method offers a good compromise between the more accurate but slow ray tracing approach and the more efficient but less accurate direct illumination method.
%
%%%%%%%%%%%%%%%%%%%%%%%%%%%%%%%%%
%
%
%%%%%%%%%%%%%%%%%%%%%%%%%%%%%%%%%
%
%%%%%%%%%%%%%%%%%%%%%%%%%%%%%%%%%
%
\begin{table}
	\begin{center}
		\begin{tabular}{|c|c|}
			\hline
			\multicolumn{2}{|c|}{\textit{Lighting Estimation Accuracy}} \\
			\hline
			\textbf{Method}                                           & \textbf{MSE}$\downarrow$ 	   \\
			\hline
			Redner~\cite{li2018differentiable}						  & \textbf{1.854e-3}	           \\
			DI~\cite{ravi2020pytorch3d} 						      & 9.512e-3              		   \\
			\hline
			Ours    									   	     	  & 2.667e-3                  	   \\
			\hline
		\end{tabular}
	\end{center}
	\caption
	{
		Lighting estimation accuracy averaged over 10 different scenes.
		Here, we evaluate the mean squared error between the rendered  and  ground truth  images.
		Our approach outperforms the direct illumination method, and is close to the ray tracing method. 
	}
	\label{tab:lighting}
\end{table}
%
%%%%%%%%%%%%%%%%%%%%%%%%%%%%%%%%%
%
%%%%%%%%%%%%%%%%%%%%%%%%%%%%%%%%%
%%%%%%%%%%%%%%%%%%%%%%%%%%%%%%%%%
%
\paragraph{6D Pose Optimization.}
Here, we optimize the 6D pose of the object with known light and texture, see Fig.~\ref{fig:Pose} and Tab.~\ref{tab:Pose} from one image.
Our renderer provides useful supervision for correctly optimizing the rigid pose which can be seen qualitatively as well as quantitatively.
Compared to previous methods, we clearly outperform DI method as they cannot use the shadow cues as signal.
In contrast, the cast shadows directly provide supervision for the unknown pose in our case.
Interestingly, we also outperform the ray tracing approach~\cite{li2018differentiable}.
We hypothesize that it is difficult for ray tracing renderers to optimize the geometry because of their local nature of computation. 
The gradients at any point in the scene are propagated through the rays which reach this point.
Thus, the gradients mostly rely on local properties of the scene. 
In contrast, the shadow computation at any point in our renderer directly depends on the global scene geometry, which leads to less noisy gradients. 
%
%%%%%%%%%%%%%%%%%%%%%%%%%%%%%%%%%
%
%
%%%%%%%%%%%%%%%%%%%%%%%%%%%%%%%%%
%
%%%%%%%%%%%%%%%%%%%%%%%%%%%%%%%%%
%
\begin{table}
	\begin{center}
		\begin{tabular}{|c|c|c|c|}
			\hline
			\multicolumn{4}{|c|}{\textit{Pose Estimation Accuracy}} \\
			\hline
			\textbf{Method}                                             & \textbf{MDE}  $\downarrow$        & \textbf{RE} $\downarrow$ 		  & \textbf{TE} $\downarrow$    \\
			\hline
			Redner~\cite{li2018differentiable}							& 12.72		                            & 0.2289           	   	           & \textbf{4.633e-4}     \\
			DI~\cite{ravi2020pytorch3d} 							    & 23.53	                                &  1.0154      		                & 7.719e-2     \\
			\hline
			Ours    									   	            & \textbf{6.420}	                        & \textbf{0.0702}   	            & 1.075e-3     \\
			\hline
		\end{tabular}
	\end{center}
	\caption
	{
		Pose reconstruction accuracy averaged over 10 different scenes.
		The mean distance error is calculated between the ground truth and  optimized vertex positions. 
		We report MDE as a percentage of the diagonal length of the bounding box around the ground truth shape.
		Rotation error (RE) is calculated as the angle between the rotations, and translation error (TE) as the squared distance between the translations.
		We outperform both approaches.
	}
	\label{tab:Pose}
\end{table}
%
%%%%%%%%%%%%%%%%%%%%%%%%%%%%%%%%%
%
\paragraph{Geometry Optimization.}
In Fig.~\ref{fig:geometry} and Tab.~\ref{tab:Geometry}, we evaluate how our renderer can be used for estimating the geometric deformation of an object, given the \textit{undeformed} template as well as the lighting and albedo of the scene.
To this end, the rotation and translation parameters of the embedded graph are optimized.
Like in the case of global pose reconstruction, our approach outperforms the previous works~\cite{li2018differentiable, ravi2020pytorch3d} both qualitatively and quantitatively.
%
%%%%%%%%%%%%%%%%%%%%%%%%%%%%%%%%%
%
%
%%%%%%%%%%%%%%%%%%%%%%%%%%%%%%%%%
%
\begin{table}
	\begin{center}
		\begin{tabular}{|c|c|}
			\hline
			\multicolumn{2}{|c|}{\textit{Geometry Reconstruction Accuracy}} \\
			\hline
			\textbf{Method}                                             & \textbf{MDE}  $\downarrow$    		  \\
			\hline
			Redner~\cite{li2018differentiable}							& 18.66		                  	   	              \\
			DI~\cite{ravi2020pytorch3d} 							    & 28.98	                           		          \\
			\hline
			Ours    									   	            & \textbf{11.80}	                    	          \\
			\hline
		\end{tabular}
	\end{center}
	\caption
	{
		Geometry reconstruction accuracy averaged over 10 different scenes.
		We outperform both approaches in terms of MDE (explained in Tab.~\ref{tab:Pose}).
	}
	\label{tab:Geometry}
\end{table}
%
%%%%%%%%%%%%%%%%%%%%%%%%%%%%%%%%%
%
%%%%%%%%%%%%%%%%%%%%%%%%%%%%%%%%%
%%%%%%%%%%%%%%%%%%%%%%%%%%%%%%%%%
%
\subsection{Runtime}
We evaluate the runtime of our approach and compare it with the state of the art.
Tab.~\ref{tab:runtime} reports the average runtime per iteration in milliseconds (ms) and frames per second (fps).
Our approach is significantly (up to two orders of magnitude) faster than the ray tracing  approach~\cite{li2018differentiable}.
We are close to the runtime of the method of Ravi~\etal~\cite{ravi2020pytorch3d} for most tasks except for pose and geometry estimation.
This shows that our shadow computation requires a minimal overhead compared to DI method, with the advantage of higher-quality reconstruction of all scene properties.
%
%%%%%%%%%%%%%%%%%%%%%%%%%%%%%%%%%
%
\begin{table}
	\begin{center}
	
		\begin{tabular}{|c|c|c|c|c|}
			\hline
			\multicolumn{5}{|c|}{\textit{Runtime Performance(in ms)$\downarrow$ }} \\
			\hline
			\textbf{Method}                                             & \textbf{Texture }      & \textbf{Light }              & 
			\textbf{Pose } 	 & 
			\textbf{Geometry }   \\
			\hline
			Redner~\cite{li2018differentiable}						& 3261  & 2938  & 4473   &4897                        \\
			DI~\cite{ravi2020pytorch3d} 							& \textbf{18}  & \textbf{18}  & \textbf{81} & \textbf{233}
			      		                    \\
			\hline
			Ours    									   	            & \textbf{18}	& \textbf{18}  & 216 &391                                          
			\\     
			\hline
		\end{tabular}
	\end{center}
	\caption
	{
		Runtime performance averaged over multiple iterations including forward and backward passes.
		Our approach clearly outperforms the ray tracing renderer, and is  close to the DI method, indicating that our shadow computation is efficient.
	}
	\label{tab:runtime}
\end{table}
%
%%%%%%%%%%%%%%%%%%%%%%%%%%%%%%%%%
%
%%%%%%%%%%%%%%%%%%%%%%%%%%%%%%%%%
%%%%%%%%%%%%%%%%%%%%%%%%%%%%%%%%%
%
\subsection{Reconstruction from Shadows}
\begin{figure}
	\includegraphics[width=\linewidth]{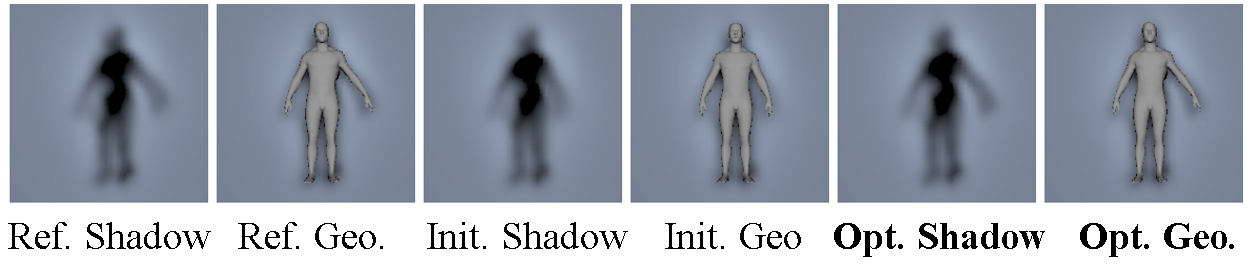} 
	\caption
	{
		Shadow fitting result.
		The shadow alone can be used to recover the geometric deformations.
	}
	\label{fig:shadow}
\end{figure}

In Fig.~\ref{fig:shadow}, we show how the shadow alone can be used to optimize the object deformations. 
We optimize the embedded graph deformation parameters and compare the rendered shadows with the reference shadow image.
This demonstrates that shadows offer very strong cues about the scenes. 
Our method, in contrast to direct illumination renderers can utilize these cues for accurate reconstruction.
%
%%%%%%%%%%%%%%%%%%%%%%%%%%%%%%%%%
%
%
%%%%%%%%%%%%%%%%%%%%%%%%%%%%%%%%%
%
%%%%%%%%%%%%%%%%%%%%%%%%%%%%%%%%%
%%%%%%%%%%%%%%%%%%%%%%%%%%%%%%%%%
%
\subsection{Ablation}
In Tab.~\ref{tab:ablation}, we study the influence of the number of spheres used to approximate the underlying geometry of the object.
We solve for the lighting reconstruction, see Fig.~\ref{fig:lightResult}.
Even a small number like 100 spheres can approximate the geometry quite well.
Adding more spheres improves the quality of reconstruction.
We also evaluate the influence of the number of spherical harmonics coeffcients while fixing the number of spheres to 100 using the same setting.
A small number of coefficients can already approximate the scene lighting quite well.
Adding more coefficients leads to better results as higher frequency lighting can be captured as well.
Importantly, for all experiments with different numbers of spheres and lighting coefficients,  we achieve a runtime of 18ms due to GPU parallelization .
%
%%%%%%%%%%%%%%%%%%%%%%%%%%%%%%%%%
%
\begin{table}
	\begin{center}
		\begin{tabular}{|c|c|c|c|c|}
			\hline
			\multicolumn{5}{|c|}{\textit{Ablation Study}} \\
			\hline
		    \textbf{Spheres}                                        & 	50 	                    & 	100 	            & 		150          & 	200            	     \\
			\textbf{MSE}$\downarrow \times 10^{-3}$                 & 1.445                     & 1.349                 & 1.336            & \textbf{1.335}                      \\

			\hline
		    \textbf{Coeffs}                                         & 16                        &  25 	               & 36            &    49                           	    \\
			\textbf{MSE}$\downarrow \times 10^{-3}$                 & 2.437                     & 1.982                 & 1.629          & \textbf{1.493}              \\
			\hline
		\end{tabular}
	\end{center}
	\caption
	{
		Ablation study.
		Even a small number of spheres and lighting coefficients can lead to plausible results.
		Adding more spheres and lighting coefficients constantly improves the result as surface details can be better approximated and higher frequency lighting can be modeled.
	}
	\label{tab:ablation}
\end{table}
%
%%%%%%%%%%%%%%%%%%%%%%%%%%%%%%%%%
%
%
\section{Discussion}
\begin{figure}
	\includegraphics[width=\linewidth]{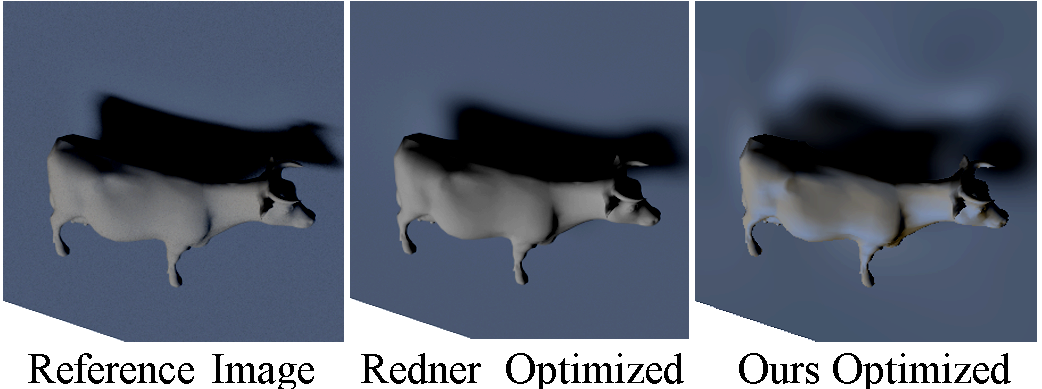} 
	\caption
	{
		Limitation.
	    Our method fails to reconstruct high-frequency lighting, such as a directional light in this example.
	}
	\label{fig:limitaion}
\end{figure}

%
%%%%%%%%%%%%%%%%%%%%%%%%%
%
High frequency lighting cannot be modelled well by our approach due to the low-dimensional SH representation.
Fig.~\ref{fig:limitaion} shows one such case with incident illumination just from a single direction. 
Our method requires the precomputation of the embedded deformation graph. 
This graph is well-suited for one object category, however, it would not be sufficient to deform very different shapes.
Our approach does not consider inter-reflectance and non-diffuse surfaces, which are interesting directions for future work. 
%
%%%%%%%%%%%%%%%%%%%%%%%%%%%%%%%%%
%
%
%%%%%%%%%%%%%%%%%%%%%%%%%
%
\section{Conclusion}
%
%%%%%%%%%%%%%%%%%%%%%%%%%
%
We proposed a method for efficient and differentiable shadow computation that can be used for various inverse problems. 
We show that our approach achieves competitive results compared to ray tracing methods at much faster runtimes.
Further, we outperform direct illumination renderers which do not model shadows.
We demonstrate that shadows are important cues in images, and take the first steps towards using efficient high-quality differentiable and shadow-aware rendering for inverse problems.
%

%%%%%%%%%%%%%%%%%%%%%%%%%%%%%%%%%%%%%

\paragraph{Acknowledgments.}
This work has been supported by the ERC Consolidator Grant 4DReply (770784), and Lise Meitner Postdoctoral Fellowship.
%
%%%%%%%%%%%%%%%%%%%%%%%%%%%%%%%%%%%%%

{\small
\bibliographystyle{ieee_fullname}
\bibliography{egbib}
}

\end{document}